\documentclass[letterpaper]{article} 
\usepackage{aaai2027}  
\usepackage[hyphens]{url}  
\usepackage{graphicx} 
\usepackage{natbib}  
\usepackage{caption} 
\usepackage{algorithm}
\usepackage{algorithmic}

\usepackage{subcaption}
\usepackage{placeins}
\usepackage{longtable}

\usepackage{newfloat}
\usepackage{listings}
\DeclareCaptionStyle{ruled}{labelfont=normalfont,labelsep=colon,strut=off} 
\floatstyle{ruled}
\newfloat{listing}{tb}{lst}{}
\floatname{listing}{Listing}

\usepackage{booktabs}
\usepackage{xcolor}

\title{DataWeave: Deploying Human–LLM Analytics for\\ Exploratory Structured Data Analysis}
\author{
    Raquib Bin Yousuf\textsuperscript{\rm 1}\equalcontrib, Harith Laxman\textsuperscript{\rm 1}\equalcontrib, Vitaliy Shkremetko\textsuperscript{\rm 1}, Eunice Son\textsuperscript{\rm 1}, Shambhavi Verma\textsuperscript{\rm 1}, \\
    Brian O'Leary\textsuperscript{\rm 2}, Venketesh Subramony\textsuperscript{\rm 2}, Sylvain Nazef\textsuperscript{\rm 2}, Jacquelyn Elias\textsuperscript{\rm 2},\\
    Ron Coddington\textsuperscript{\rm 2},
    Chris Contakes\textsuperscript{\rm 2}, Michael Riley\textsuperscript{\rm 2}, Naren Ramakrishnan\textsuperscript{\rm 1}
}
\affiliations{
    \textsuperscript{\rm 1}Virginia Tech, VA, USA.
    \textsuperscript{\rm 2}The Chronicle of Higher Education, Washington, DC, USA \\
    raquib@vt.edu, naren@cs.vt.edu
}

\begin{document}

\maketitle

\begin{abstract}
Data journalism, the practice of using data analysis to surface newsworthy stories, depends increasingly on the ability of reporters and investigative journalists to uncover trends, disparities, and accountability narratives. In practice, exploring large structured datasets remains slow and brittle: journalists must navigate hundreds of variables across many datasets over years, understand data coding conventions, and write non-trivial analysis code while hypotheses evolve. Although LLMs are often touted as ``ask in English, get SQL/answers,'' real newsroom workflows expose recurring failures, e.g., schema mismatches and drift, misread domain semantics and units, and silent assumptions. We present DataWeave, a system that addresses these needs by combining conversational interaction, schema grounding, analytical planning, and executable query generation to support exploratory analysis over structured data. Rather than treating LLMs as autonomous answer engines, DataWeave frames them as interactive partners whose outputs can be inspected, corrected, and steered as hypotheses shift. We present a case study with professional journalists using our system to analyze the U.S. Department of Education's Integrated Postsecondary Education Data System (IPEDS), a high-stakes public dataset with substantial domain semantics and frequent schema updates. We also report how deployment experience and iterative refinement shaped the current DataWeave architecture and its analytical workflow. Our findings distill design principles and deployment lessons for trustworthy human--LLM collaboration in structured data analysis.
\end{abstract}


\section{Introduction}
Large-scale structured datasets play a central role in decision making in education~\cite{hanushek2005does, dynarski2015missing}, public policy~\cite{einav2014data, jarmin2016big}, and journalism ~\cite{kasica2023dirty, hao2024design, hamilton2009accountability}. Despite their importance, these datasets remain difficult to explore in practice ~\cite{milo2020automating, kandel2012enterprise, liu2019understanding}. Analysts 
must navigate hundreds of variables spread across multiple tables, interpret cryptic schema documentation, and write non-trivial code as questions evolve~\cite{han2023data, chai2023demystifying}. This friction disproportionately affects domain experts, such as journalists, who possess deep contextual knowledge but may lack the technical expertise or time for low-level data wrangling~\cite{kandel2011wrangler, chattopadhyay2020s}.

Recent advances in large language models (LLMs) have renewed interest in conversational interfaces for data analysis ~\cite{barbon2024large, drosos2024s, chandel2022training, lai2023ds}. Many existing systems cast structured data reasoning as a question--answering task, where users issue a natural language query and the model returns a summarized result ~\cite{gomez2024automatic, chopra2023conversational}. Other approaches pursue fully autonomous analytical agents that plan, execute, and refine analyses without user intervention ~\cite{hu2025agentgen}. While promising in controlled settings, these paradigms are poorly aligned with real-world exploratory workflows ~\cite{xie2024waitgpt, wang2018mismatch, feng2023xnli, ferdowsi2023coldeco, narechania2021diy}. In practice, analysts rarely know the \emph{right question} in advance ~\cite{gu2024analysts, jiang2022discovering, xu2022ide}, and fully autonomous execution obscures intermediate assumptions, making it difficult to trust or correct results ~\cite{khurana2024and, ning2024insights, li2024dawn, hu2024large, olausson2023self}.

\begin{figure*}[ht]
    \centering
    \includegraphics[width=0.9\linewidth]{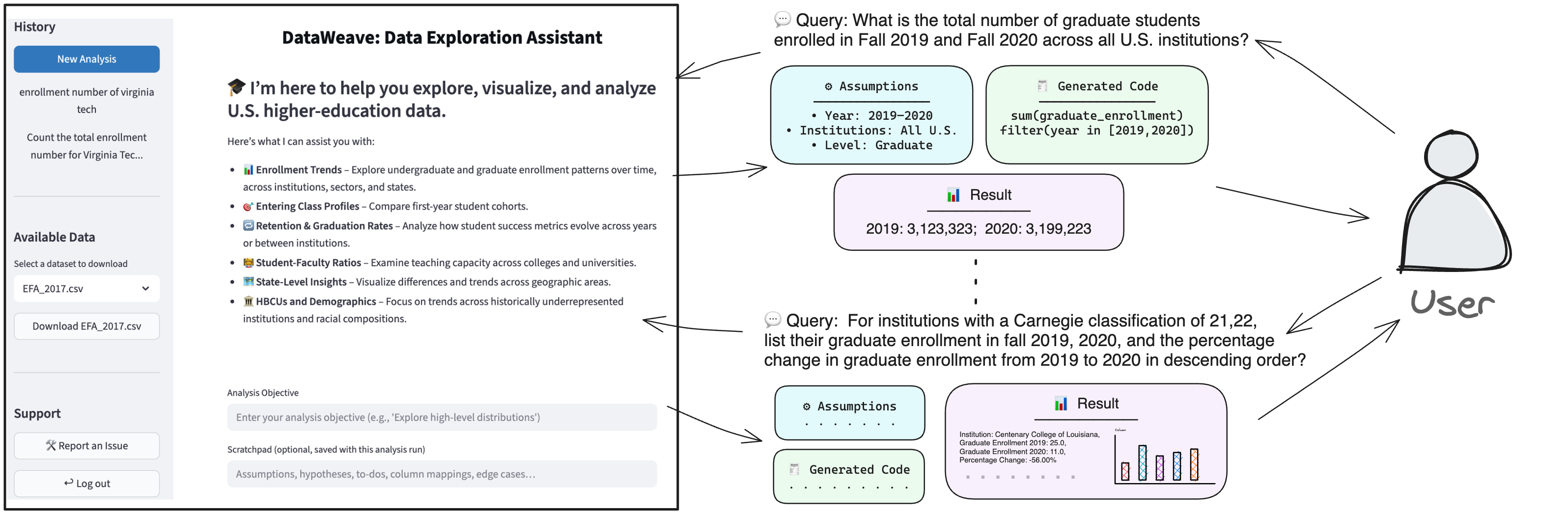}
    \caption{Illustration of the DataWeave interaction workflow. A user poses a natural-language query over higher-education data, from which DataWeave constructs inspectable analytical artifacts including analytical assumptions, executable queries, and results (tables or visualizations). These artifacts are surfaced to the user to support verification, correction, and iterative refinement.
    }
    \label{fig:conceptual_arch}
\end{figure*}

Our work is motivated by a year-long collaboration with education journalists analyzing the U.S.\ Department of Education’s Integrated Postsecondary Education Data System (IPEDS). IPEDS is a large, heterogeneous collection of tabular datasets covering enrollment, demographics, finances, and outcomes across thousands of U.S.\ 
educational institutions. Although publicly available, IPEDS poses significant challenges: schemas vary across years and tables, column semantics are highly domain-specific, and missing or inconsistently reported values are common. Journalists working with IPEDS emphasized the need not for automated answers, but for tools that support \emph{iterative sensemaking}, including the ability to explore trends, test hypotheses, inspect intermediate results, and correct errors as understanding evolves.

Through early prototypes, we observed that fully autonomous LLM-driven analytics often failed in subtle but consequential ways. Examples included assuming the presence of nonexistent columns, misinterpreting table semantics, or silently producing misleading aggregates. Crucially, these errors were frequently easy for domain experts to detect and fix once exposed. This observation led to a key insight: for structured data analysis, \emph{human guidance is not a limitation, but a strength}. Small, well-timed user interventions can substantially improve analytical correctness, interpretability, and trust.

Based on these findings, we present \textbf{DataWeave}, an interactive, human-in-the-loop system for exploratory analysis of structured and numerical data using LLMs. DataWeave combines intent interpretation, schema grounding, analytical planning, query formulation, and bounded execution within an agent-mediated conversational workflow, while users retain control over analytical goals and interpretation. Users articulate open-ended analytical goals, while the system generates schema-grounded executable queries and preserves intermediate analytical traces for inspection and refinement.
DataWeave supports cross-table joins, longitudinal comparisons, and subgroup analyses common in data journalism workflows.
Figure~\ref{fig:conceptual_arch} shows the interactive analysis workflow.

The current architecture emerged through continued deployment and refinement, where experience with an earlier stage-wise implementation informed the final organization of these analytical functions.
Our contributions are:
\begin{enumerate}

\item We introduce \textbf{DataWeave}, a deployed, human-in-the-loop analytics system for use by professional journalists.
As part of this study, we identify recurring failure modes in LLM-based structured data analysis, e.g., schema drift, silent aggregation errors, and misaligned temporal assumptions, and show how exposing intermediate artifacts makes them readily correctable.

\item We distill a set of transferable design principles for human–LLM collaboration in high-stakes analytical workflows, including clarification before assumption, objective-conditioned schema grounding, inspectable executable queries, bounded recovery, and guided agent delegation.

\item We report deployment experience from professional newsroom use and show how iterative refinement of the system led to the current agent-mediated architecture while preserving its core analytical workflow.

\item Through deployment analysis and a retrospective IPEDS case study, we demonstrate how these design choices support iterative sensemaking while preserving editorial control and accountability.

\end{enumerate}

\begin{figure*}[ht]
    \centering
    \includegraphics[width=0.93\linewidth]{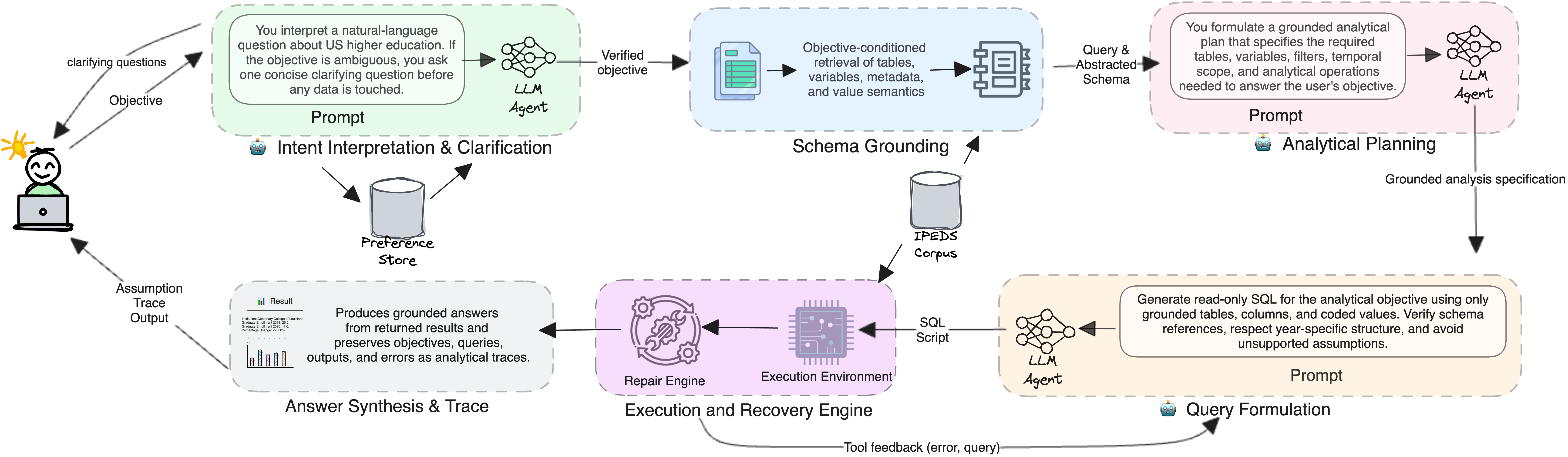}
    \caption{DataWeave architecture. User objectives are interpreted and clarified, grounded through objective-conditioned schema retrieval, translated into analytical plans and read-only SQL, and executed with bounded recovery. Queries, results, and errors are preserved as analytical traces to support transparency and reproducibility.}
    \label{fig:chron_arch}
\end{figure*}

\begin{table*}[h]
\centering
\resizebox{0.75\textwidth}{!}{
\begin{tabular}{@{}p{3cm}p{5cm}p{8cm}@{}}
\toprule
\textbf{Design Principle} & \textbf{What this means} & \textbf{How this is realized in DataWeave} \\ \hline
Permission Over Forgiveness &
Ambiguous analytical intent should be resolved before unsupported assumptions propagate into the analysis. &
DataWeave requests clarification when key information such as scope, terminology, or temporal context is ambiguous, while preserving the user's analytical objective. \\ \hline
Executable Queries as Inspectable Analytical Artifacts &
Executable queries and intermediate results should remain inspectable rather than being hidden behind natural-language answers. &
DataWeave grounds analytical responses in executable SQL and preserves queries, returned results, and provenance as part of the analytical trace. \\ \hline
Objective-Conditioned Schema Grounding &
LLM reasoning should use compact schema and semantic context relevant to the current analytical objective. &
DataWeave retrieves relevant tables, variables, descriptions, and coded-value semantics on demand rather than exposing the complete schema to the model. \\
\hline
Error Exposure Over Silent Robustness &
Analytical systems should favor explicit error exposure
and recoverable failure states over silent corrections that may alter
analytical intent. &
DataWeave treats execution errors
as informative signals rather than exceptional conditions to be
hidden. Failures are surfaced directly to the user, and any automated
recovery is deliberately bounded and constrained. \\ \hline
Guided Agent Delegation &
LLMs may coordinate analytical steps dynamically, while user intent and interpretation remain under human control. &
DataWeave allows the agent to sequence schema discovery, planning, query formulation, and recovery while using clarification and inspectable traces to preserve user oversight. \\
\bottomrule
\end{tabular}
}
\caption{Condensed view of our design principles, what they mean, and how DataWeave realizes them.}
\label{tab:principles-condensed}
\end{table*}

\section{Related Work}
Recent research has examined how LLMs can assist with structured data analysis via natural language interfaces, program synthesis, and visualization support~\cite{wu2024automated, liu2024ava, liu2023jarvix}. OpenTab~\cite{kong2024opentab}, for example, studies open-domain table reasoning where models must first identify relevant tables and then generate SQL and answers. InsightBench~\cite{sahu2024insightbench} proposes a benchmark for evaluating agents that autonomously generate multi-step insights from curated datasets. Other systems emphasize interaction and visualization ~\cite{shen2022towards}, such as LIDA~\cite{dibia2023lida} and Chat2VIS~\cite{maddigan2023chat2vis, shen2022towards}, which convert user requests into charts and lightweight exploratory analyses~\cite{tian2024chartgpt}. Mixed-initiative tools like Data Formulator and Data Formulator 2~\cite{wang2023data_formulator,wang2025data_formulator_2} support iterative specification of data transformations and visual encodings. Code-oriented assistants including GPT-4 \cite{achiam2023gpt, baktash2023gpt, rodrigues2025gpt, stribling2024model} as a Data Analyst~\cite{cheng2023gpt, huang2024evaluating, li2024comparing} and Data-Copilot~\cite{zhang2023data_copilot} help users generate and execute analysis code, while agent frameworks such as AutoGen~\cite{wu2024autogen} investigate multi-agent coordination for complex tasks~\cite{dorri2018multi}.

A closely related line of work is Text-to-SQL, which studies how to translate natural language questions into executable SQL over relational databases~\cite{luo2025natural}. Recent advances have improved schema linking~\cite{wang2020rat}, prompt design, and execution-guided generation~\cite{liu2025survey, li2024dawn, yu2019cosql, xie2025opensearch, scholak2021picard}. These systems typically assume a well-posed question and a relatively stable schema~\cite{li2024dawn}, and aim to produce a correct query for execution.

DataWeave overlaps with prior work in using LLMs to generate executable analytical artifacts, but targets high-stakes exploratory analysis over heterogeneous, evolving public datasets. In this setting, questions are underspecified, schemas drift across releases, and analysts refine goals iteratively. Accordingly, DataWeave emphasizes objective-conditioned schema grounding, analytical planning, constrained query execution, and persistent traces that record assumptions, errors, and revisions.
Rather than optimizing for autonomous insight generation or rapid visualization, DataWeave is designed for collaborative investigative workflows where human oversight is central and intermediate steps remain visible and contestable.

\section{System Architecture}
\label{sec:architecture}

At a high level, DataWeave
combines LLM-based intent interpretation and analytical planning with schema-grounded query formulation and deterministic execution over structured IPEDS data.
User interactions are coordinated through an agent-mediated workflow comprising intent interpretation and clarification, schema grounding, analytical planning, query formulation, execution and recovery, and answer synthesis with persistent analytical traces.
Figure~\ref{fig:chron_arch} shows the core system modules and their roles.

\subsection{Front-End and Session Management}

DataWeave provides a conversational web interface that maintains a persistent analysis session per authenticated user. The interface shows the conversation, clarification requests, analytical activity, and returned results, with tools to browse prior logs and manage sessions. Persistent state maintains conversation history and the analytical context required to support follow-up questions, while per-user storage preserves prior turns, outputs, errors, and user-approved preferences.
Authentication and per-user storage support isolated, long-running analytical workflows.

\subsection{Data and Metadata Representation}
DataWeave operates over multiple table families spanning time periods. Each IPEDS collection year is represented through year-specific relational tables together with metadata catalogs describing tables, variables, data types, and categorical value sets. Upon ingestion, table and column names are normalized to reduce brittleness in query formulation.

Rather than assuming a fixed global schema, DataWeave treats schema information as year-specific metadata that may vary across releases. This representation allows downstream components to reason about table and variable selection, categorical value semantics, and temporal alignment as the schema changes over time.
Instead of placing the complete schema in the model context, DataWeave retrieves relevant table, variable, and value-set metadata on demand and uses this information as structured grounding for subsequent analysis.

\subsection{Intent Interpretation and Clarification}
Given a user turn, DataWeave interprets the analytical objective and determines whether sufficient information is available to proceed. If the objective contains an ambiguous abbreviation, missing temporal scope, or other information required for analysis, the system requests a concise clarification before querying the data. The clarified objective then provides the basis for schema grounding and subsequent analysis.

\subsection{Schema Grounding}
Given the clarified analytical objective, DataWeave retrieves relevant information from the year-specific metadata catalog. The system first identifies candidate tables and then retrieves descriptions of relevant variables and, when needed, the allowed values and meanings of categorical codes. This objective-conditioned retrieval provides a compact grounded schema context without requiring the complete database schema to be included in the model prompt.

\subsection{Analytical Planning}
Given the analytical objective and grounded schema context, DataWeave forms an analytical plan specifying the tables, variables, filters, temporal scope, and analytical operations required to answer the question. Planning is grounded in retrieved schema and semantic information, including the meanings of coded variables, rather than relying on model memory. The resulting analytical specification constrains subsequent query formulation.

\subsection{Query Formulation}
Based on the analytical plan, DataWeave formulates one or more read-only SQL queries that retrieve, filter, join, and aggregate the required data. Query formulation is constrained by the grounded schema and domain-specific rules, including year-specific table structure, variable semantics, and categorical codes. Table and variable names are therefore derived from retrieved metadata rather than assumed by the model.

\subsection{Execution and Recovery}
Generated queries are executed against a read-only data store, with execution restricted to retrieval operations and bounded result sets. Execution can produce either concrete query results or explicit errors. When a query fails, the execution feedback is returned to the analytical workflow, allowing the query to be reformulated using the original objective, grounded schema context, and observed error. Recovery remains bounded rather than allowing uncontrolled execution. Queries, returned results, and errors are retained as part of the analytical trace.

\subsection{Answer Synthesis and Trace Persistence}
DataWeave synthesizes the final response from the results returned through successful query execution rather than from unsupported model assumptions. The response retains relevant provenance, such as the collection year and source tables, while the underlying interaction is preserved as an analytical trace containing the user objective, tool interactions, queries, returned results, and execution errors.

These traces are persisted across sessions, enabling users to inspect prior analytical steps, continue an investigation through follow-up questions, and resume earlier exploratory workflows. Persistent conversational state also allows clarification and other user interventions to remain part of the analytical process.

\subsection{Design Principles behind DataWeave}
\label{sec:principles}

The process of deploying, evaluating, and iteratively refining DataWeave has led us to articulate key
design principles as shown
in Table~\ref{tab:principles-condensed}.
We highlight them here as potential takeaways for the design of reliable human-LLM analytics systems for complex, heterogeneous structured datasets.

\section{Deployment Experience and System Evolution}
\label{sec:deployment}
\begin{table}[!h]
\centering
\small
\begin{tabular}{lr}
\toprule
\textbf{Metric} & \textbf{Value} \\
\midrule
\multicolumn{2}{l}{\textit{Scale}} \\
Users & 18 \\
Sessions & 298 \\
Analytical steps & 797 \\
Avg.\ steps/session & 2.67 \\
\midrule
\multicolumn{2}{l}{\textit{Session Depth}} \\
Multi-step sessions & 48.0\% \\
Sessions $\geq$5 steps & 12.8\% \\
Max steps in a session & 35 \\
\midrule
\multicolumn{2}{l}{\textit{Engagement}} \\
Median sessions/user & 8 \\
Max sessions/user & 105 \\
Active users ($\geq$5 sessions) & 12 \\
\midrule
\multicolumn{2}{l}{\textit{Iterative Behavior}} \\
Refinement transitions & 66.7\% \\
\bottomrule
\end{tabular}
\caption{Summary of deployment usage.}
\label{tab:deployment-summary}
\end{table}

\begin{table}[!t]
\centering
\small
\begin{tabular}{lcc}
\toprule
\textbf{Metric} & \textbf{Mean} & \textbf{Median} \\
\midrule
Columns inferred per step & 8.94 & 6 \\
Columns used per step & 6.03 & 5 \\
Column overlap & 5.38 & 5 \\
Overlap ratio (used covered by inferred) & 0.86 & 1.00 \\
Overlap ratio (inferred utilized) & 0.64 & 0.67 \\
Used-only columns (not inferred) & 0.65 & 0 \\
Inferred-only columns (not used) & 3.56 & 2 \\
\bottomrule
\end{tabular}
\caption{Alignment between planner-inferred columns and columns actually used in execution (deployment logs). Higher overlap indicates stronger schema grounding.}
\label{tab:planner_alignment}
\end{table}
DataWeave was deployed for use by professional journalists and data analysts at \emph{The Chronicle of Higher Education}. We collected interaction logs to characterize how the system was used in practice and how its design evolved through deployment. The initial deployed version used explicit stage-wise orchestration, while subsequent refinements led to the agent-mediated architecture described in Section~\ref{sec:architecture}.

\subsection{Deployment and Iterative Use}
Deployment logs contain 298 analysis sessions comprising 797 analytical steps from 18 users (Table~\ref{tab:deployment-summary}). Usage extended beyond isolated interactions: the median user completed eight sessions, and 12 of the 18 users completed at least five sessions. Sessions frequently extended beyond a single query and included follow-up questions and refinements of earlier analyses.

\begin{figure}[h]
\centering
\begin{subfigure}{\linewidth}
    \centering
    \includegraphics[width=0.5\linewidth]{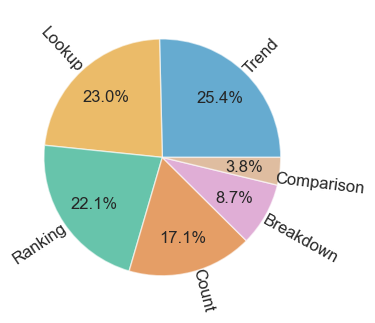}
    \caption{Primary intents (DAI).}
    \label{fig:primary}
\end{subfigure}
\begin{subfigure}{\linewidth}
    \centering
    \includegraphics[width=.98\linewidth]{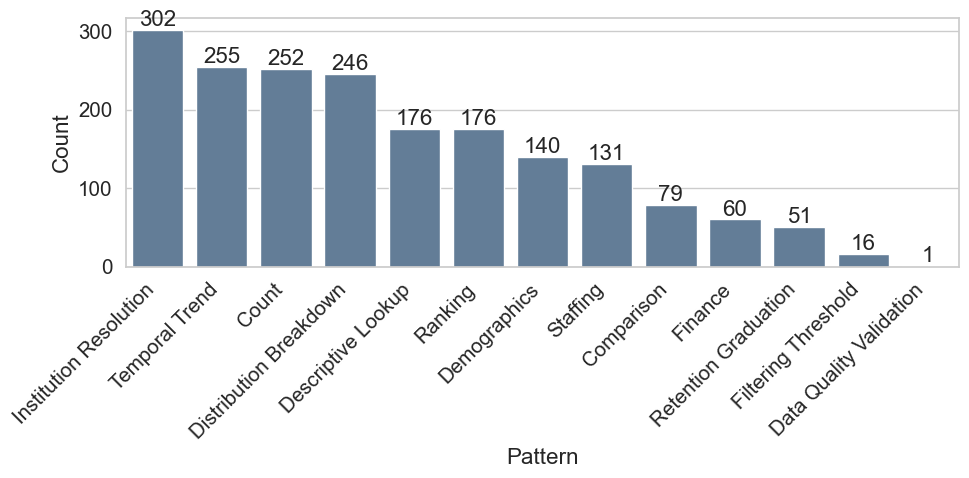}
    \caption{Analytical dimension tags (ADT).}
    \label{fig:tag}
\end{subfigure}
\caption{Distribution of analytical patterns across real-world analytical steps. (a) Each step is assigned one dominant analytical intent (Data quality validation with $n=1$ not shown). (b) Steps may also receive multiple analytical dimension tags (ADT) capturing additional analytical properties.}
\label{fig:patterns}
\end{figure}

To characterize these workflows, each analytical step was assigned one \emph{Dominant Analytical Intent} (DAI), describing its primary purpose, and one or more \emph{Analytical Dimension Tags} (ADT), describing secondary properties such as comparison or demographic focus (Figure~\ref{fig:patterns}). A substantial fraction of transitions between steps refined a previous objective, indicating that DataWeave was commonly used as an iterative analytical workspace rather than a single-turn question-answering system.

\subsection{Schema Grounding and Planning Behavior}
\begin{figure}[h]
\centering
\includegraphics[width=0.9\linewidth]{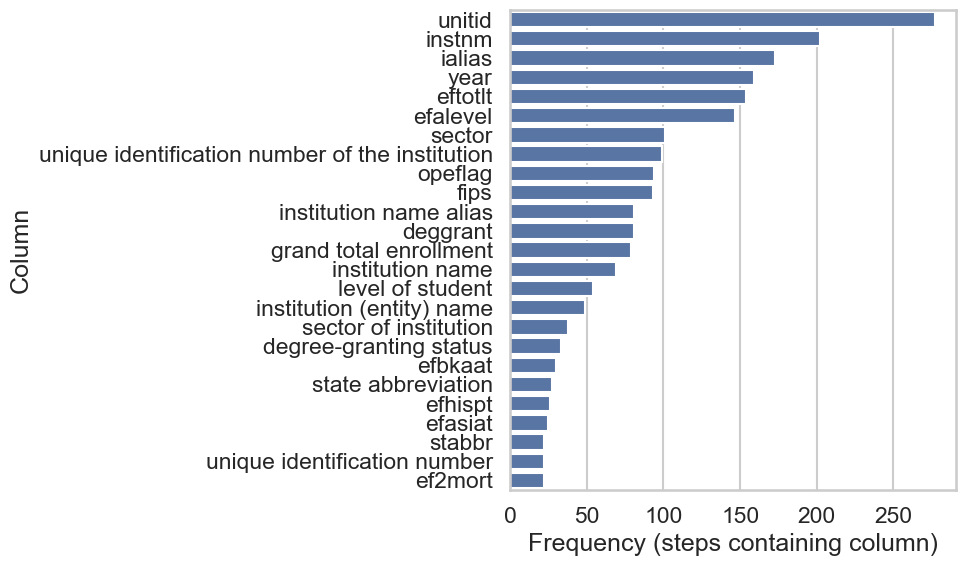}
\caption{Most frequently used columns across real-world analytical steps. Core identifiers (e.g., \texttt{unitid}, \texttt{instnm}) and enrollment-related variables dominate usage, reflecting consistent grounding in entity resolution and enrollment analysis.}
\label{fig:used_columns}
\end{figure}

Deployment logs also allow us to examine how the system selected and used the underlying IPEDS schema. Core identifiers such as \texttt{unitid} and \texttt{instnm}, together with frequently used enrollment variables such as \texttt{efalevel} and \texttt{eftotlt}, appeared most often in executed analyses (Figure~\ref{fig:used_columns}).

\begin{figure}[h]
\centering
\includegraphics[width=0.85\linewidth]{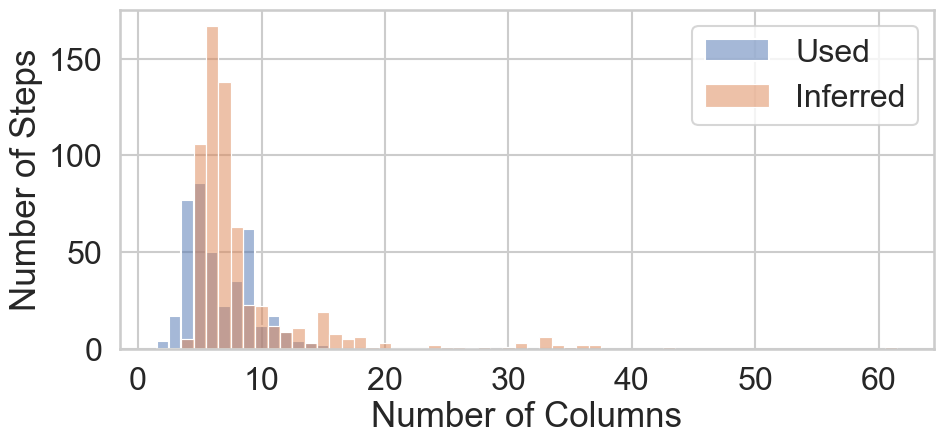}
\caption{Distribution of planner-inferred versus executed columns per analytical step. The planner typically proposes a broader column set, while execution uses a smaller subset. This pattern indicates conservative schema grounding followed by selective use during execution.}
\label{fig:planned_vs_used}
\end{figure}

\begin{figure}[h]
\centering
\includegraphics[width=0.70\linewidth]{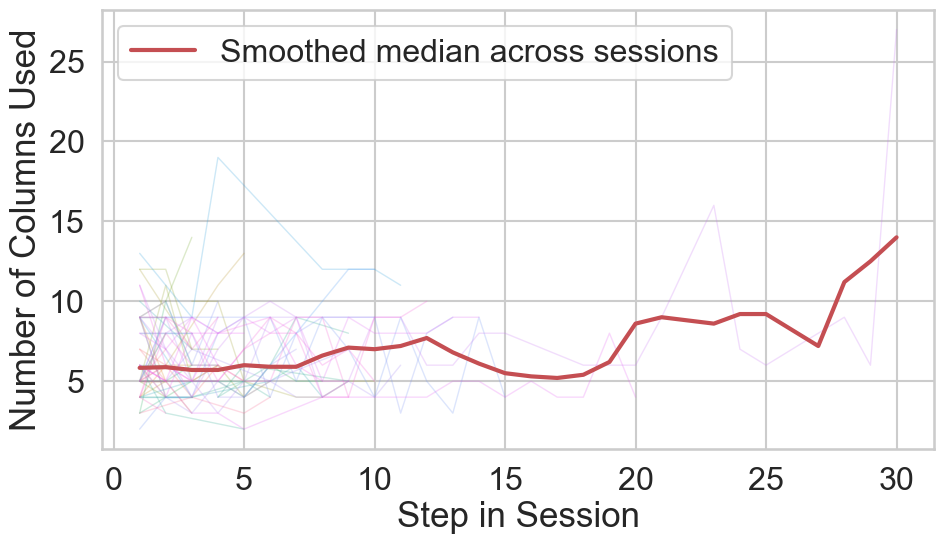}
\caption{Column footprint across step progression within sessions. Thin lines show individual sessions and the bold line shows the median. Later steps often involve broader column usage, reflecting refinement and deeper investigation.}
\label{fig:step_progression}
\end{figure}

The analytical planner generally identified a broader set of candidate columns than were ultimately used during execution (Table~\ref{tab:planner_alignment} and Figure~\ref{fig:planned_vs_used}). This pattern suggests conservative schema coverage followed by task-specific selection. Column usage also tended to increase in later steps of longer sessions (Figure~\ref{fig:step_progression}), consistent with investigations becoming more detailed as users refined their questions.

\subsection{From Staged to Agent-Mediated Orchestration}
\label{sec:extending_deployment}
The initial deployed version organized intent handling, schema grounding, planning, code generation, and execution as explicit application-defined stages. As DataWeave evolved, these core functions were retained, but their coordination shifted toward ReAct-style agents that dynamically sequence schema discovery, planning, query formulation, and execution feedback.

The final architecture therefore preserves the original analytical structure while allowing the execution path to adapt more directly to the task.

\begin{figure}[!t]
\centering
\begin{subfigure}{\linewidth}
    \centering
    \includegraphics[width=0.9\linewidth, trim={0 0.7cm 0 0}, clip]{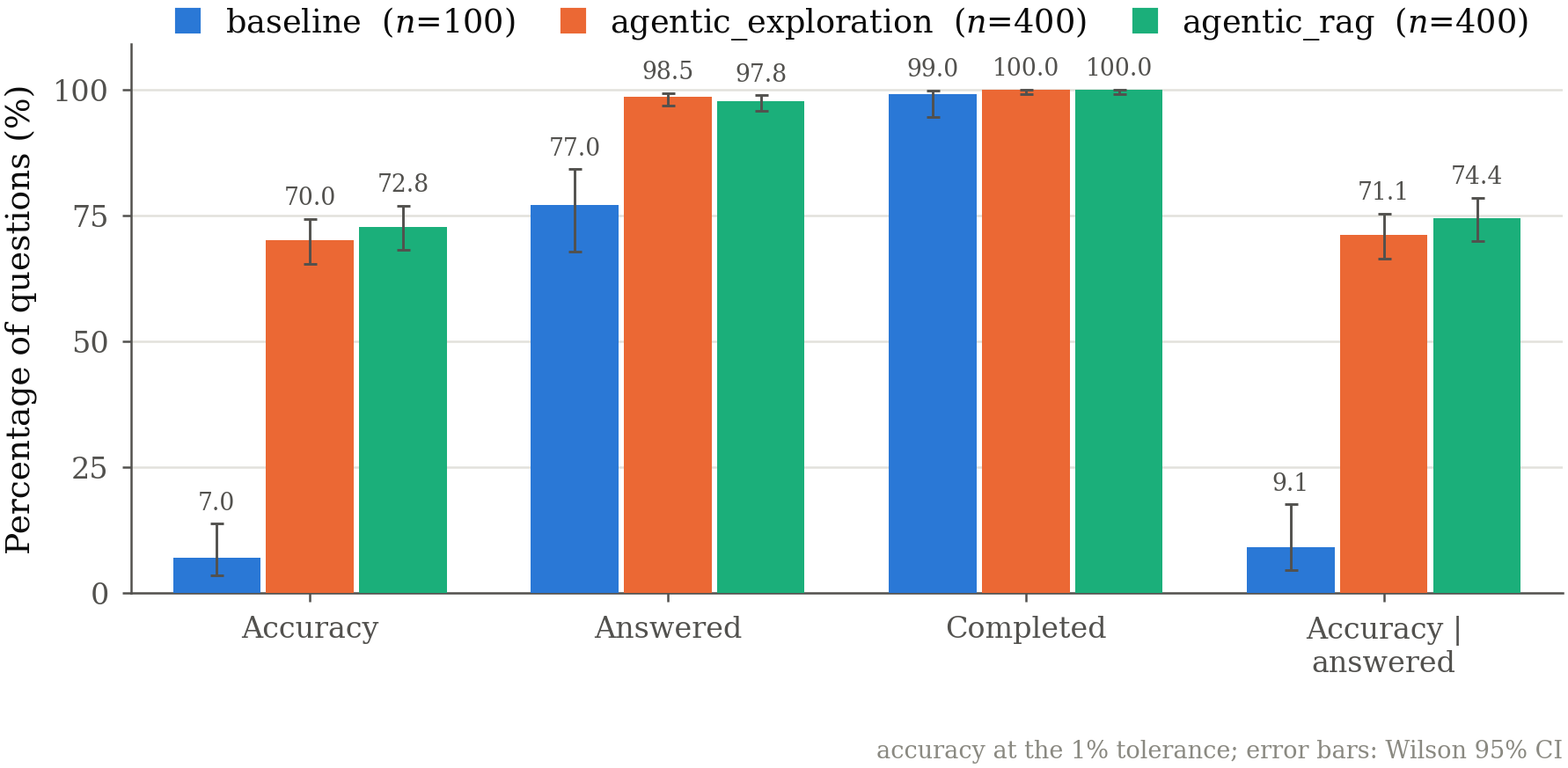}
    \caption{Headline evaluation metrics at the 1\% error tolerance.}
    \label{fig:headline_metrics}
\end{subfigure}
\begin{subfigure}{\linewidth}
    \centering
    \includegraphics[width=0.90\linewidth, trim={0 1.3cm 0 0}, clip]{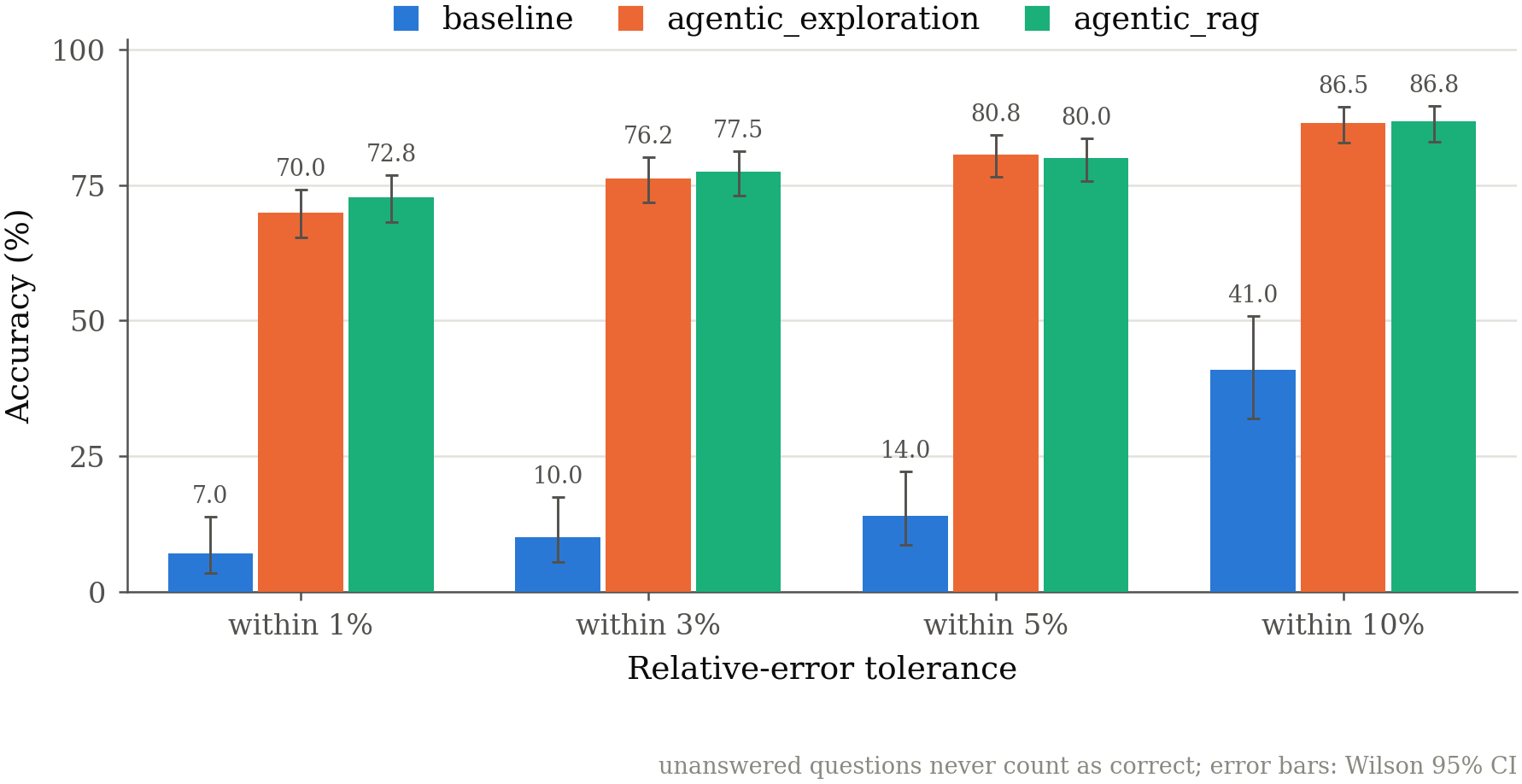}
    \caption{Accuracy across relative-error tolerances.}
    \label{fig:accuracy_tolerance}
\end{subfigure}
\caption{Answer correctness across the evaluated DataWeave configurations. The earlier prototype baseline is evaluated on a 100-question subset covering the five most recent years, while the agentic configurations are evaluated on the full 400-question benchmark. Error bars show Wilson 95\% confidence intervals.}
\label{fig:rq1_results}
\end{figure}

\begin{figure}[!t]
\centering
\begin{subfigure}{\linewidth}
    \centering
    \includegraphics[width=0.9\linewidth, trim={0 1cm 0 0}, clip]{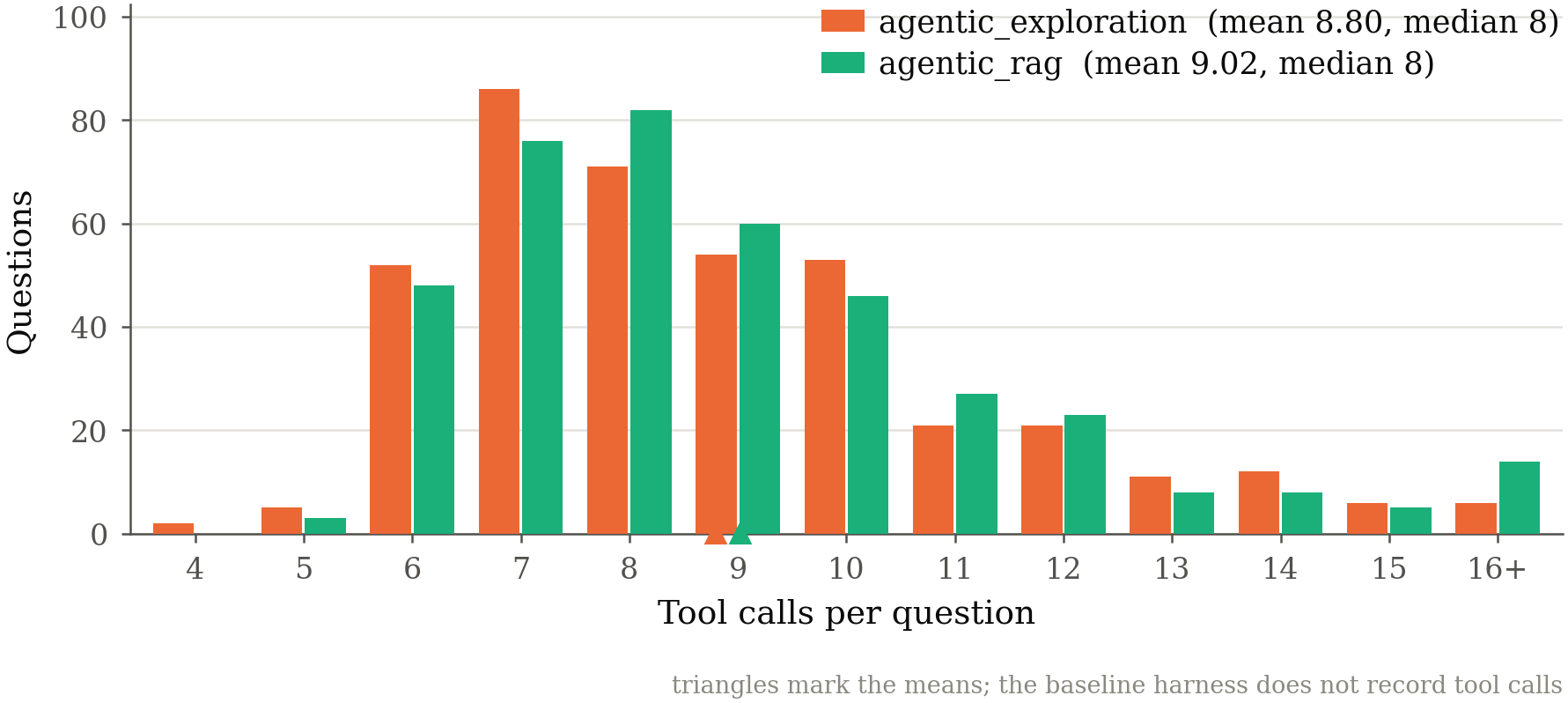}
    \caption{Distribution of tool calls per question.}
    \label{fig:tool_calls}
\end{subfigure}
\begin{subfigure}{\linewidth}
    \centering
    \includegraphics[width=0.8\linewidth, trim={0 0 0 2cm}, clip]{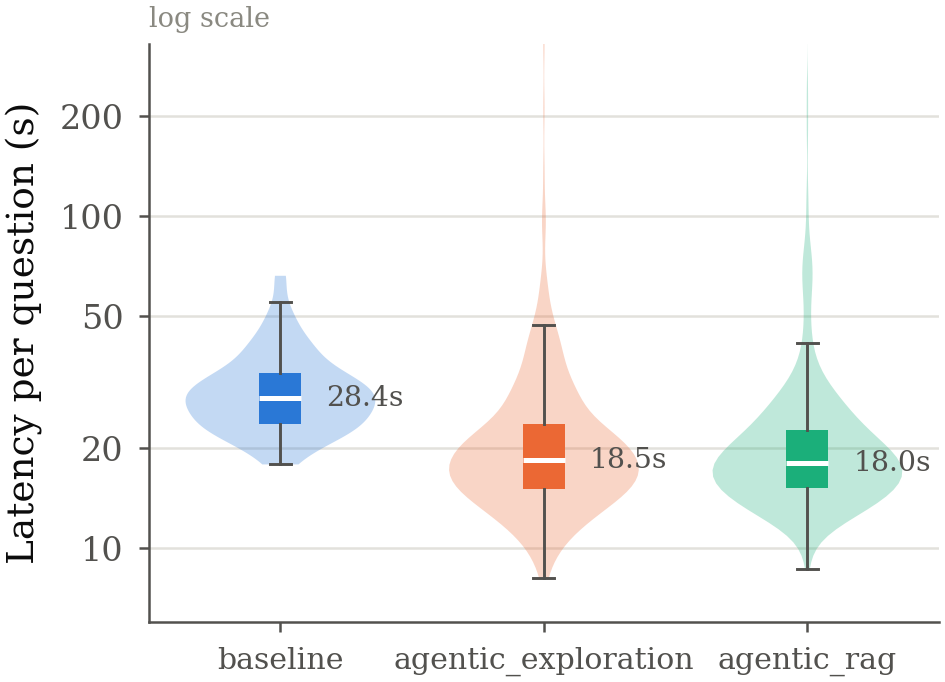}
    \caption{Latency per question.}
    \label{fig:latency}
\end{subfigure}
\begin{subfigure}{\linewidth}
    \centering
    \includegraphics[width=0.8\linewidth, trim={0 0.8cm 9.4cm 2cm}, clip]{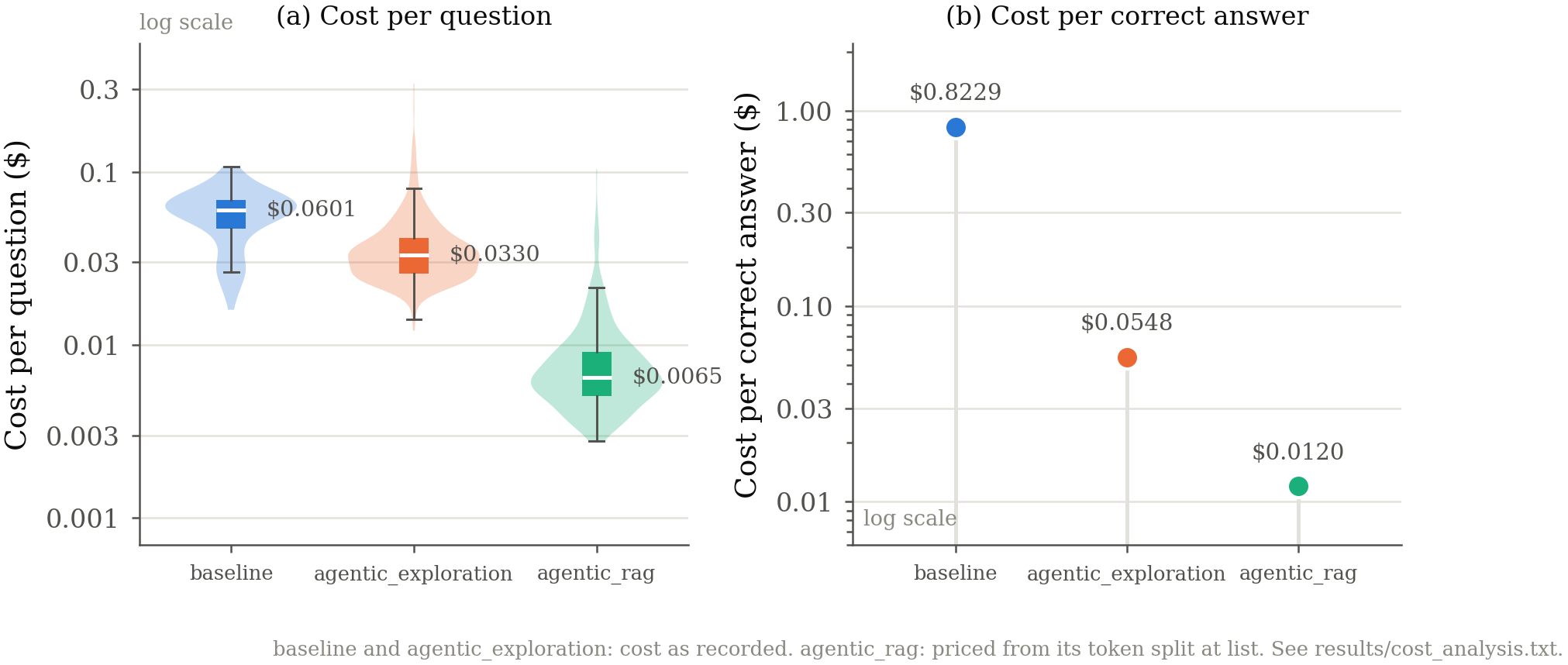}
    \caption{Cost per question.}
    \label{fig:cost}
\end{subfigure}
\caption{Operational behavior of the evaluated configurations. Tool-call statistics are available for the agentic configurations; the earlier prototype harness does not record tool calls. Latency is reported for all evaluated runs.}
\label{fig:operational_behavior}
\end{figure}

\begin{figure*}[!t]
    \centering
    \includegraphics[width=0.76\linewidth]{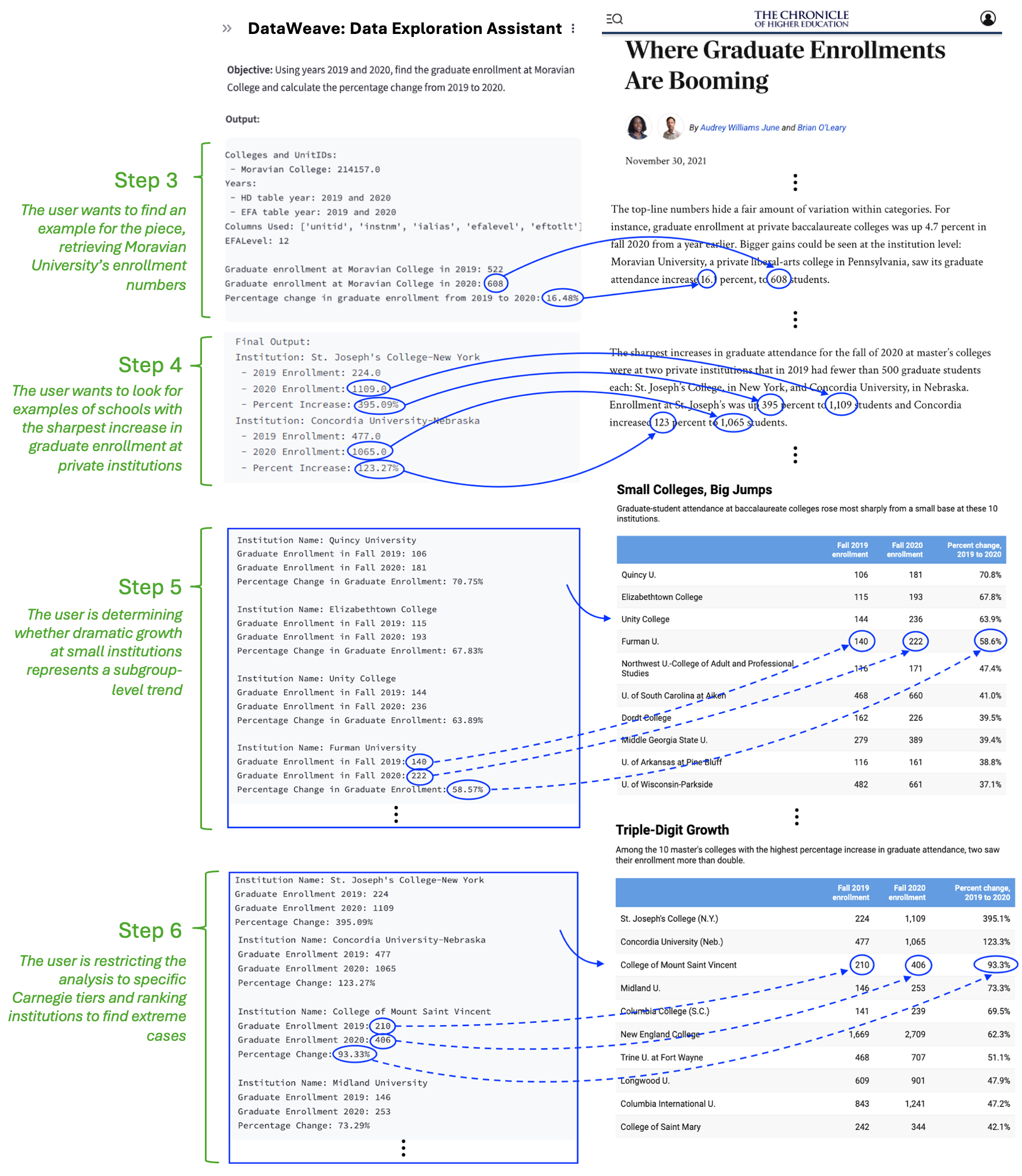}
    \caption{Retrospective case study aligning DataWeave analyses with a 2021 \emph{Chronicle of Higher Education} article on graduate enrollment trends. The figure shows selected steps in a reconstructed analytical workflow, where DataWeave queries reproduce reported statistics. Arrows link computed outputs to corresponding claims and tables in the article, illustrating how conversational analysis can support journalistic verification and story development.}
    \label{fig:usecase}
\end{figure*}

\subsection{Operational Implications}
\label{sec:agent_tradeoffs}
Agent-mediated orchestration increased analytical flexibility, but also made schema exploration and computational cost more dependent on the agent's reasoning path. This reinforced the need for structured schema access, explicit planning, constrained execution, and persistent traces around the agent workflow. The evolution also showed that analytical flexibility and integration compatibility are separate deployment considerations.

\section{Evaluation}
\label{sec:evaluation}
The deployment analysis in the previous section characterizes how DataWeave is used in open-ended newsroom workflows. Here, we evaluate the system on structured questions with published reference answers and examine both answer quality and operational behavior.

We structure the evaluation around three questions:
\begin{itemize}
    \item \textbf{RQ1: Answer Correctness.} How accurately does DataWeave answer structured IPEDS questions with known reference values?
    \item \textbf{RQ2: Operational Behavior.} How do the evaluated system configurations differ in tool use, latency, and computational cost?
    \item \textbf{RQ3: Investigative Utility.} Can DataWeave reproduce and support analytical trajectories underlying real journalistic reporting?
\end{itemize}

RQ1--RQ2 use a benchmark derived from published IPEDS Trend Generator questions, while RQ3 reconstructs the analysis behind a previously published article.

\subsection{Evaluation Setup}
\label{sec:eval_setup}
We evaluate DataWeave on 400 questions derived from 45 published IPEDS Trend Generator questions spanning 12 subject areas and 20 collection years (2004--05 through 2023--24). Questions include direct totals, institution-filtered totals, and categorical breakdowns, and are sampled with a fixed seed across years and question types.

Each template is rewritten once into natural language while preserving the requested measure, year, and analytical dimension. Rewrites are programmatically checked, with the original template used as fallback when needed.

Because published Trend Generator values assume active, Title IV-participating institutions in the 50 states and the District of Columbia, this scope is supplied through an evaluation-only prompt suffix. The deployed system prompt and tools are otherwise unchanged.

The two agentic configurations are evaluated on all 400 questions. The earlier prototype baseline is evaluated on 100 questions from the five most recent years because its execution design requires the relevant year-specific tables to be loaded into memory.

Accuracy is measured within 1\% relative error, or within 0.5 percentage points for percentage-valued answers. We also report answer rate, completion rate, latency, tool calls, and cost.

\subsection{RQ1: Answer Correctness}
\label{sec:rq1}
As shown in Figure~\ref{fig:headline_metrics}, the baseline achieves 7.0\% accuracy, compared with 70.0\% for Agentic Exploration and 72.8\% for Agentic RAG. Both agentic configurations answer nearly all questions and complete all runs.

Figure~\ref{fig:accuracy_tolerance} shows that this gap remains across broader error tolerances. The two agentic configurations perform similarly, suggesting that retrieval-based grounding does not materially change answer correctness on this benchmark.

\subsection{RQ2: Operational Behavior}
\label{sec:rq2}

We next examine the execution behavior of the agentic configurations. Agentic Exploration and Agentic RAG use a similar number of tool calls, with mean values of 8.80 and 9.02, respectively, and a median of eight calls for both (Figure~\ref{fig:operational_behavior}). This suggests that retrieval-based grounding does not substantially change the number of analytical interactions required per question.

Latency is also similar across the two configurations, with median values of 18.5~s for Agentic Exploration and 18.0~s for Agentic RAG. The earlier prototype shows a higher median latency of 28.4~s on its evaluated subset.

Operational cost differs more substantially. Mean cost per question is \$0.0601 for the baseline, \$0.0330 for Agentic Exploration, and \$0.0065 for Agentic RAG. Costs reflect the accounting available from each evaluation backend and are therefore interpreted as operational rather than strictly normalized costs.

\subsection{RQ3: Investigative Utility in Story Construction}
To answer RQ3, we evaluate whether DataWeave can support investigative workflows comparable to real journalistic practice. We do so through a retrospective case study based on a previously published 2021 \emph{Chronicle of Higher Education} article on graduate enrollment trends. The article predates DataWeave and was produced independently by journalists. Our goal is not to generate the article, but to reconstruct the analytical workflow underlying its claims and assess whether DataWeave can reproduce the reported statistics and support similar lines of inquiry.

As shown in Figure~\ref{fig:usecase},
story construction is typically sequential: reporters detect a signal, segment it by subgroup, validate sector-level claims, and identify institution-level examples.
In this example, beginning with
baseline U.S. graduate enrollment trends,
the story narrows down into
particular sectors, surfacing
outliers as story angles,
followed by subgroup validation.

This retrospective reconstruction shows that DataWeave can reproduce key reported statistics and support the sequential analytical moves observed in real reporting. The system functions as a query-to-story scaffold rather than a single-turn answer engine. By enabling signal detection, segmentation, validation, and outlier discovery with inspectable computations, DataWeave supports investigative sensemaking and verification.
\section{Lessons Learned}
Our deployment of DataWeave with professional journalists yielded the following lessons for human–LLM structured data analysis.

\noindent
{\bf Collaboration Better Fits Investigative Workflows.}
Users treated DataWeave as a collaborative partner rather than an autonomous analyst. The system supported analysis and query generation, while journalists retained control over interpretation and decision making. This suggests that inspectable and steerable workflows are better suited to exploratory, high-stakes analysis than end-to-end autonomy.

\noindent
{\bf Transparency supports Trust and Accountability.}
Journalists regularly inspected generated queries and intermediate outputs before accepting results, reflecting professional norms around verification and defensibility. By making analytical steps explicit, DataWeave supported verification practices central to data journalism.
Similar demands arise in policy, public health, and education, where analyses must be explainable, revisitable, and defensible.

\noindent
{\bf Agent Delegation Shifts Analytical Complexity.}
Agent-mediated orchestration increases flexibility, but shifts more complexity into the model's reasoning path. This can produce broader and less predictable schema exploration, reinforcing the need for structured grounding, constrained execution, and persistent traces.

\noindent
{\bf Deployment Fit Extends Beyond Analytical Capability.}
A system may provide the required analytical capability but still differ from how an organization expects to integrate it into existing workflows. Our deployment experience showed that reasoning flexibility, efficiency, and integration compatibility should be treated as separate design considerations.

\section{Conclusion}
Through deployment with professional education journalists, we showed that conversational, query-centric workflows can support iterative sensemaking while preserving transparency and user control. Deployment experience shaped DataWeave's current architecture, combining schema grounding, analytical planning, agent-mediated coordination, and constrained execution. The results further show that these design choices can substantially improve correctness while maintaining practical latency and cost. More broadly, reliable LLM analytics should balance agent flexibility with structured grounding, human oversight, and deployment compatibility. Future work will extend DataWeave to additional domains and explore further forms of guided autonomy while preserving human control.
\bibliography{aaai2027}

@article{kong2024opentab,
  title={Opentab: Advancing large language models as open-domain table reasoners},
  author={Kong, Kezhi and Zhang, Jiani and Shen, Zhengyuan and Srinivasan, Balasubramaniam and Lei, Chuan and Faloutsos, Christos and Rangwala, Huzefa and Karypis, George},
  journal={arXiv preprint arXiv:2402.14361},
  year={2024}
}

@article{sahu2024insightbench,
  title={Insightbench: Evaluating business analytics agents through multi-step insight generation},
  author={Sahu, Gaurav and Puri, Abhay and Rodriguez, Juan and Abaskohi, Amirhossein and Chegini, Mohammad and Drouin, Alexandre and Taslakian, Perouz and Zantedeschi, Valentina and Lacoste, Alexandre and Vazquez, David and others},
  journal={arXiv preprint arXiv:2407.06423},
  year={2024}
}

@article{li2024dawn,
  title={The dawn of natural language to SQL: Are we fully ready?},
  author={Li, Boyan and Luo, Yuyu and Chai, Chengliang and Li, Guoliang and Tang, Nan},
  journal={arXiv preprint arXiv:2406.01265},
  year={2024}
}

@article{liu2025survey,
  title={A Survey of Text-to-SQL in the Era of LLMs: Where are we, and where are we going?},
  author={Liu, Xinyu and Shen, Shuyu and Li, Boyan and Ma, Peixian and Jiang, Runzhi and Zhang, Yuxin and Fan, Ju and Li, Guoliang and Tang, Nan and Luo, Yuyu},
  journal={IEEE Transactions on Knowledge and Data Engineering},
  year={2025},
  publisher={IEEE}
}

@article{dibia2023lida,
  title={LIDA: A tool for automatic generation of grammar-agnostic visualizations and infographics using large language models},
  author={Dibia, Victor},
  journal={arXiv preprint arXiv:2303.02927},
  year={2023}
}

@article{maddigan2023chat2vis,
  title={Chat2vis: Generating data visualizations via natural language using ChatGPT, codex and gpt-3 large language models},
  author={Maddigan, Paula and Susnjak, Teo},
  journal={Ieee Access},
  volume={11},
  pages={45181--45193},
  year={2023},
  publisher={Ieee}
}

@article{wang2023data_formulator,
  title={Data formulator: Ai-powered concept-driven visualization authoring},
  author={Wang, Chenglong and Thompson, John and Lee, Bongshin},
  journal={IEEE Transactions on Visualization and Computer Graphics},
  volume={30},
  number={1},
  pages={1128--1138},
  year={2023},
  publisher={IEEE}
}

@inproceedings{wang2025data_formulator_2,
  title={Data Formulator 2: Iterative Creation of Data Visualizations, with AI Transforming Data Along the Way},
  author={Wang, Chenglong and Lee, Bongshin and Drucker, Steven M and Marshall, Dan and Gao, Jianfeng},
  booktitle={Proceedings of the 2025 CHI Conference on Human Factors in Computing Systems},
  pages={1--17},
  year={2025}
}

@article{cheng2023gpt,
  title={Is GPT-4 a good data analyst?},
  author={Cheng, Liying and Li, Xingxuan and Bing, Lidong},
  journal={arXiv preprint arXiv:2305.15038},
  year={2023}
}

@article{zhang2023data_copilot,
  title={Data-copilot: Bridging billions of data and humans with autonomous workflow},
  author={Zhang, Wenqi and Shen, Yongliang and Lu, Weiming and Zhuang, Yueting},
  journal={arXiv preprint arXiv:2306.07209},
  year={2023}
}

@inproceedings{wu2024autogen,
  title={Autogen: Enabling next-gen LLM applications via multi-agent conversations},
  author={Wu, Qingyun and Bansal, Gagan and Zhang, Jieyu and Wu, Yiran and Li, Beibin and Zhu, Erkang and Jiang, Li and Zhang, Xiaoyun and Zhang, Shaokun and Liu, Jiale and others},
  booktitle={First Conference on Language Modeling},
  year={2024}
}

@article{achiam2023gpt,
  title={GPT-4 technical report},
  author={Achiam, Josh and Adler, Steven and Agarwal, Sandhini and Ahmad, Lama and Akkaya, Ilge and Aleman, Florencia Leoni and Almeida, Diogo and Altenschmidt, Janko and Altman, Sam and Anadkat, Shyamal and others},
  journal={arXiv preprint arXiv:2303.08774},
  year={2023}
}

@inproceedings{khurana2024and,
  title={Why and when llm-based assistants can go wrong: Investigating the effectiveness of prompt-based interactions for software help-seeking},
  author={Khurana, Anjali and Subramonyam, Hariharan and Chilana, Parmit K},
  booktitle={Proceedings of the 29th International Conference on Intelligent User Interfaces},
  pages={288--303},
  year={2024}
}

@article{ning2024insights,
  title={Insights into natural language database query errors: From attention misalignment to user handling strategies},
  author={Ning, Zheng and Tian, Yuan and Zhang, Zheng and Zhang, Tianyi and Li, Toby Jia-Jun},
  journal={ACM Transactions on Interactive Intelligent Systems},
  volume={14},
  number={4},
  pages={1--32},
  year={2024},
  publisher={ACM New York, NY}
}

@inproceedings{kasica2023dirty,
  title={Dirty data in the newsroom: comparing data preparation in journalism and data science},
  author={Kasica, Stephen and Berret, Charles and Munzner, Tamara},
  booktitle={Proceedings of the 2023 CHI Conference on Human Factors in Computing Systems},
  pages={1--18},
  year={2023}
}

@article{han2023data,
  title={A data-driven analysis of behaviors in data curation processes},
  author={Han, Lei and Chen, Tianwa and Demartini, Gianluca and Indulska, Marta and Sadiq, Shazia},
  journal={ACM Transactions on Information Systems},
  volume={41},
  number={3},
  pages={1--35},
  year={2023},
  publisher={ACM New York, NY}
}

@inproceedings{barbon2024large,
  title={Are large language models the new interface for data pipelines?},
  author={Barbon Junior, Sylvio and Ceravolo, Paolo and Groppe, Sven and Jarrar, Mustafa and Maghool, Samira and S{\`e}des, Florence and Sahri, Soror and Van Keulen, Maurice},
  booktitle={Proceedings of the International Workshop on Big Data in Emergent Distributed Environments},
  pages={1--6},
  year={2024}
}

@inproceedings{milo2020automating,
  title={Automating exploratory data analysis via machine learning: An overview},
  author={Milo, Tova and Somech, Amit},
  booktitle={Proceedings of the 2020 ACM SIGMOD international conference on management of data},
  pages={2617--2622},
  year={2020}
}

@inproceedings{gu2024analysts,
  title={How do analysts understand and verify ai-assisted data analyses?},
  author={Gu, Ken and Shang, Ruoxi and Althoff, Tim and Wang, Chenglong and Drucker, Steven M},
  booktitle={Proceedings of the 2024 CHI Conference on Human Factors in Computing Systems},
  pages={1--22},
  year={2024}
}

@inproceedings{hao2024design,
  title={Design patterns for data-driven news articles},
  author={Hao, Shan and Wang, Zezhong and Bach, Benjamin and Pschetz, Larissa},
  booktitle={Proceedings of the 2024 CHI Conference on Human Factors in Computing Systems},
  pages={1--16},
  year={2024}
}

@inproceedings{hu2024large,
  title={Are large language models capable of causal reasoning for sensing data analysis?},
  author={Hu, Zhizhang and Zhang, Yue and Rossi, Ryan and Yu, Tong and Kim, Sungchul and Pan, Shijia},
  booktitle={Proceedings of the Workshop on Edge and Mobile Foundation Models},
  pages={24--29},
  year={2024}
}

@inproceedings{xie2024waitgpt,
  title={Waitgpt: Monitoring and steering conversational llm agent in data analysis with on-the-fly code visualization},
  author={Xie, Liwenhan and Zheng, Chengbo and Xia, Haijun and Qu, Huamin and Zhu-Tian, Chen},
  booktitle={Proceedings of the 37th Annual ACM Symposium on User Interface Software and Technology},
  pages={1--14},
  year={2024}
}

@inproceedings{drosos2024s,
  title={" It's like a rubber duck that talks back": Understanding Generative AI-Assisted Data Analysis Workflows through a Participatory Prompting Study},
  author={Drosos, Ian and Sarkar, Advait and Xu, Xiaotong and Negreanu, Carina and Rintel, Sean and Tankelevitch, Lev},
  booktitle={Proceedings of the 3rd Annual Meeting of the Symposium on Human-Computer Interaction for Work},
  pages={1--21},
  year={2024}
}

@inproceedings{hu2025agentgen,
  title={Agentgen: Enhancing planning abilities for large language model based agent via environment and task generation},
  author={Hu, Mengkang and Zhao, Pu and Xu, Can and Sun, Qingfeng and Lou, Jian-Guang and Lin, Qingwei and Luo, Ping and Rajmohan, Saravan},
  booktitle={Proceedings of the 31st ACM SIGKDD Conference on Knowledge Discovery and Data Mining V. 1},
  pages={496--507},
  year={2025}
}

@inproceedings{kandel2011wrangler,
  title={Wrangler: Interactive visual specification of data transformation scripts},
  author={Kandel, Sean and Paepcke, Andreas and Hellerstein, Joseph and Heer, Jeffrey},
  booktitle={Proceedings of the sigchi conference on human factors in computing systems},
  pages={3363--3372},
  year={2011}
}

@inproceedings{gomez2024automatic,
  title={Automatic generation of conversational interfaces for tabular data analysis},
  author={Gomez-Vazquez, Marcos and Cabot, Jordi and Claris{\'o}, Robert},
  booktitle={Proceedings of the 6th ACM Conference on Conversational User Interfaces},
  pages={1--6},
  year={2024}
}

@inproceedings{chai2023demystifying,
  title={Demystifying artificial intelligence for data preparation},
  author={Chai, Chengliang and Tang, Nan and Fan, Ju and Luo, Yuyu},
  booktitle={Companion of the 2023 International Conference on Management of Data},
  pages={13--20},
  year={2023}
}

@article{chandel2022training,
  title={Training and evaluating a jupyter notebook data science assistant},
  author={Chandel, Shubham and Clement, Colin B and Serrato, Guillermo and Sundaresan, Neel},
  journal={arXiv preprint arXiv:2201.12901},
  year={2022}
}

@inproceedings{lai2023ds,
  title={DS-1000: A natural and reliable benchmark for data science code generation},
  author={Lai, Yuhang and Li, Chengxi and Wang, Yiming and Zhang, Tianyi and Zhong, Ruiqi and Zettlemoyer, Luke and Yih, Wen-tau and Fried, Daniel and Wang, Sida and Yu, Tao},
  booktitle={International Conference on Machine Learning},
  pages={18319--18345},
  year={2023},
  organization={PMLR}
}

@inproceedings{chattopadhyay2020s,
  title={What's wrong with computational notebooks? Pain points, needs, and design opportunities},
  author={Chattopadhyay, Souti and Prasad, Ishita and Henley, Austin Z and Sarma, Anita and Barik, Titus},
  booktitle={Proceedings of the 2020 CHI conference on human factors in computing systems},
  pages={1--12},
  year={2020}
}

@article{kandel2012enterprise,
  title={Enterprise data analysis and visualization: An interview study},
  author={Kandel, Sean and Paepcke, Andreas and Hellerstein, Joseph M and Heer, Jeffrey},
  journal={IEEE transactions on visualization and computer graphics},
  volume={18},
  number={12},
  pages={2917--2926},
  year={2012},
  publisher={IEEE}
}

@article{liu2019understanding,
  title={Understanding the role of alternatives in data analysis practices},
  author={Liu, Jiali and Boukhelifa, Nadia and Eagan, James R},
  journal={IEEE transactions on visualization and computer graphics},
  volume={26},
  number={1},
  pages={66--76},
  year={2019},
  publisher={IEEE}
}

@inproceedings{jiang2022discovering,
  title={Discovering the syntax and strategies of natural language programming with generative language models},
  author={Jiang, Ellen and Toh, Edwin and Molina, Alejandra and Olson, Kristen and Kayacik, Claire and Donsbach, Aaron and Cai, Carrie J and Terry, Michael},
  booktitle={Proceedings of the 2022 CHI Conference on Human Factors in Computing Systems},
  pages={1--19},
  year={2022}
}

@article{xu2022ide,
  title={In-ide code generation from natural language: Promise and challenges},
  author={Xu, Frank F and Vasilescu, Bogdan and Neubig, Graham},
  journal={ACM Transactions on Software Engineering and Methodology (TOSEM)},
  volume={31},
  number={2},
  pages={1--47},
  year={2022},
  publisher={ACM New York, NY}
}

@inproceedings{yu2019cosql,
  title={CoSQL: A conversational text-to-SQL challenge towards cross-domain natural language interfaces to databases},
  author={Yu, Tao and Zhang, Rui and Er, Heyang and Li, Suyi and Xue, Eric and Pang, Bo and Lin, Xi Victoria and Tan, Yi Chern and Shi, Tianze and Li, Zihan and others},
  booktitle={Proceedings of the 2019 conference on empirical methods in natural language processing and the 9th international joint conference on natural language processing (EMNLP-IJCNLP)},
  pages={1962--1979},
  year={2019}
}

@article{xie2025opensearch,
  title={Opensearch-SQL: Enhancing text-to-SQL with dynamic few-shot and consistency alignment},
  author={Xie, Xiangjin and Xu, Guangwei and Zhao, Lingyan and Guo, Ruijie},
  journal={Proceedings of the ACM on Management of Data},
  volume={3},
  number={3},
  pages={1--24},
  year={2025},
  publisher={ACM New York, NY, USA}
}

@article{scholak2021picard,
  title={PICARD: Parsing incrementally for constrained auto-regressive decoding from language models},
  author={Scholak, Torsten and Schucher, Nathan and Bahdanau, Dzmitry},
  journal={arXiv preprint arXiv:2109.05093},
  year={2021}
}

@article{chopra2023conversational,
  title={Conversational challenges in ai-powered data science: Obstacles, needs, and design opportunities},
  author={Chopra, Bhavya and Singha, Ananya and Fariha, Anna and Gulwani, Sumit and Parnin, Chris and Tiwari, Ashish and Henley, Austin Z},
  journal={arXiv preprint arXiv:2310.16164},
  year={2023}
}

@article{olausson2023self,
  title={Is self-repair a silver bullet for code generation?},
  author={Olausson, Theo X and Inala, Jeevana Priya and Wang, Chenglong and Gao, Jianfeng and Solar-Lezama, Armando},
  journal={arXiv preprint arXiv:2306.09896},
  year={2023}
}

@article{shen2022towards,
  title={Towards natural language interfaces for data visualization: A survey},
  author={Shen, Leixian and Shen, Enya and Luo, Yuyu and Yang, Xiaocong and Hu, Xuming and Zhang, Xiongshuai and Tai, Zhiwei and Wang, Jianmin},
  journal={IEEE transactions on visualization and computer graphics},
  volume={29},
  number={6},
  pages={3121--3144},
  year={2022},
  publisher={IEEE}
}

@article{baktash2023gpt,
  title={Gpt-4: A review on advancements and opportunities in natural language processing},
  author={Baktash, Jawid Ahmad and Dawodi, Mursal},
  journal={arXiv preprint arXiv:2305.03195},
  year={2023}
}

@article{rodrigues2025gpt,
  title={Is GPT-4 Fair? An Empirical Analysis in Automatic Short Answer Grading},
  author={Rodrigues, Luiz and Xavier, Cleon and Costa, Newarney and Gasevic, Dragan and Mello, Rafael Ferreira},
  journal={Computers and Education: Artificial Intelligence},
  pages={100428},
  year={2025},
  publisher={Elsevier}
}

@article{stribling2024model,
  title={The model student: GPT-4 performance on graduate biomedical science exams},
  author={Stribling, Daniel and Xia, Yuxing and Amer, Maha K and Graim, Kiley S and Mulligan, Connie J and Renne, Rolf},
  journal={Scientific Reports},
  volume={14},
  number={1},
  pages={5670},
  year={2024},
  publisher={Nature Publishing Group UK London}
}

@article{huang2024evaluating,
  title={Evaluating ChatGPT-4.0’s data analytic proficiency in epidemiological studies: A comparative analysis with SAS, SPSS, and R},
  author={Huang, Yeen and Wu, Ruipeng and He, Juntao and Xiang, Yingping},
  journal={Journal of Global Health},
  volume={14},
  pages={04070},
  year={2024}
}

@article{li2024comparing,
  title={Comparing GPT-4 and human researchers in health care data analysis: qualitative description study},
  author={Li, Kevin Danis and Fernandez, Adrian M and Schwartz, Rachel and Rios, Natalie and Carlisle, Marvin Nathaniel and Amend, Gregory M and Patel, Hiren V and Breyer, Benjamin N},
  journal={Journal of Medical Internet Research},
  volume={26},
  pages={e56500},
  year={2024},
  publisher={JMIR Publications Toronto, Canada}
}

@article{wu2024automated,
  title={Automated data visualization from natural language via large language models: An exploratory study},
  author={Wu, Yang and Wan, Yao and Zhang, Hongyu and Sui, Yulei and Wei, Wucai and Zhao, Wei and Xu, Guandong and Jin, Hai},
  journal={Proceedings of the ACM on Management of Data},
  volume={2},
  number={3},
  pages={1--28},
  year={2024},
  publisher={ACM New York, NY, USA}
}

@article{dorri2018multi,
  title={Multi-agent systems: A survey},
  author={Dorri, Ali and Kanhere, Salil S and Jurdak, Raja},
  journal={IEEE Access},
  volume={6},
  pages={28573--28593},
  year={2018},
  publisher={IEEE}
}

@inproceedings{wang2020rat,
  title={RAT-SQL: Relation-aware schema encoding and linking for text-to-sql parsers},
  author={Wang, Bailin and Shin, Richard and Liu, Xiaodong and Polozov, Oleksandr and Richardson, Matthew},
  booktitle={Proceedings of the 58th annual meeting of the association for computational linguistics},
  pages={7567--7578},
  year={2020}
}

@article{luo2025natural,
  title={Natural language to SQL: State of the art and open problems},
  author={Luo, Yuyu and Li, Guoliang and Fan, Ju and Chai, Chengliang and Tang, Nan},
  journal={Proceedings of the VLDB Endowment},
  volume={18},
  number={12},
  pages={5466--5471},
  year={2025},
  publisher={VLDB Endowment}
}

@article{tian2024chartgpt,
  title={Chartgpt: Leveraging llms to generate charts from abstract natural language},
  author={Tian, Yuan and Cui, Weiwei and Deng, Dazhen and Yi, Xinjing and Yang, Yurun and Zhang, Haidong and Wu, Yingcai},
  journal={IEEE Transactions on Visualization and Computer Graphics},
  volume={31},
  number={3},
  pages={1731--1745},
  year={2024},
  publisher={IEEE}
}

@article{einav2014data,
  title={The data revolution and economic analysis},
  author={Einav, Liran and Levin, Jonathan},
  journal={Innovation policy and the economy},
  volume={14},
  number={1},
  pages={1--24},
  year={2014},
  publisher={University of Chicago Press Chicago, IL}
}

@article{jarmin2016big,
  title={Big data and the transformation of public policy analysis},
  author={Jarmin, Ron S and O'Hara, Amy B},
  journal={Journal of Policy Analysis and Management},
  volume={35},
  number={3},
  pages={715--721},
  year={2016},
  publisher={JSTOR}
}

@inproceedings{hamilton2009accountability,
  title={Accountability Through Algorithm},
  author={Hamilton, James T and Turner, Fred},
  booktitle={Center for Advanced Study in the Behavioral Sciences Summer Workshop. Available online: http://web. stanford. edu/\~{}$\{$$\}$ fturner/Hamilton\% 20Turner\% 20Acc\% 20by\% 20Alg\% 20Final. pdf (accessed on 5 February 2021)},
  year={2009}
}

@article{hanushek2005does,
  title={Does school accountability lead to improved student performance?},
  author={Hanushek, Eric A and Raymond, Margaret E},
  journal={Journal of Policy Analysis and Management: The Journal of the Association for Public Policy Analysis and Management},
  volume={24},
  number={2},
  pages={297--327},
  year={2005},
  publisher={Wiley Online Library}
}

@article{dynarski2015missing,
  title={The missing manual: Using National Student Clearinghouse data to track postsecondary outcomes},
  author={Dynarski, Susan M and Hemelt, Steven W and Hyman, Joshua M},
  journal={Educational Evaluation and Policy Analysis},
  volume={37},
  number={1\_suppl},
  pages={53S--79S},
  year={2015},
  publisher={Sage Publications Sage CA: Los Angeles, CA}
}

@inproceedings{wang2018mismatch,
  title={Mismatch of expectations: How modern learning resources fail conversational programmers},
  author={Wang, April Y and Mitts, Ryan and Guo, Philip J and Chilana, Parmit K},
  booktitle={Proceedings of the 2018 CHI Conference on Human Factors in Computing Systems},
  pages={1--13},
  year={2018}
}

@inproceedings{liu2024ava,
  title={AVA: towards autonomous visualization agents through visual perception-driven decision-making},
  author={Liu, Shusen and Miao, Haichao and Li, Zhimin and Olson, Matthew and Pascucci, Valerio and Bremer, P-T},
  booktitle={Computer Graphics Forum},
  volume={43},
  number={3},
  pages={e15093},
  year={2024},
  organization={Wiley Online Library}
}

@inproceedings{liu2023jarvix,
  title={JarviX: A LLM no code platform for tabular data analysis and optimization},
  author={Liu, Shang-Ching and Wang, ShengKun and Chang, Tsungyao and Lin, Wenqi and Hsiung, Chung-Wei and Hsieh, Yi-Chen and Cheng, Yu-Ping and Luo, Sian-Hong and Zhang, Jianwei},
  booktitle={Proceedings of the 2023 Conference on Empirical Methods in Natural Language Processing: Industry Track},
  pages={622--630},
  year={2023}
}

@article{feng2023xnli,
  title={Xnli: Explaining and diagnosing nli-based visual data analysis},
  author={Feng, Yingchaojie and Wang, Xingbo and Pan, Bo and Wong, Kam Kwai and Ren, Yi and Liu, Shi and Yan, Zihan and Ma, Yuxin and Qu, Huamin and Chen, Wei},
  journal={IEEE Transactions on Visualization and Computer Graphics},
  volume={30},
  number={7},
  pages={3813--3827},
  year={2023},
  publisher={IEEE}
}

@inproceedings{ferdowsi2023coldeco,
  title={ColDeco: An end user spreadsheet inspection tool for AI-generated code},
  author={Ferdowsi, Kasra and Williams, Jack and Drosos, Ian and Gordon, Andrew D and Negreanu, Carina and Polikarpova, Nadia and Sarkar, Advait and Zorn, Benjamin},
  booktitle={2023 IEEE Symposium on Visual Languages and Human-Centric Computing (VL/HCC)},
  pages={82--91},
  year={2023},
  organization={IEEE}
}

@inproceedings{narechania2021diy,
  title={DIY: Assessing the correctness of natural language to SQL systems},
  author={Narechania, Arpit and Fourney, Adam and Lee, Bongshin and Ramos, Gonzalo},
  booktitle={Proceedings of the 26th International Conference on Intelligent User Interfaces},
  pages={597--607},
  year={2021}
}
\end{document}